\documentclass[10pt]{article}

\usepackage{graphicx}
\usepackage{subfig}
\usepackage{sidecap}
\usepackage{amsfonts}
\usepackage{amsmath}
\usepackage{amssymb}
\usepackage{amsthm}

\usepackage{array}
\usepackage{float}
\usepackage{multirow}

\usepackage{psfrag}
\usepackage{graphics}
\usepackage{float}

\usepackage{algorithm}
\usepackage{algpseudocode}
\usepackage{epstopdf}

\usepackage{soul}
\usepackage{xspace}
\usepackage{soul,color}

\usepackage{hyperref}
\hypersetup{colorlinks=true}

\usepackage[margin=1in]{geometry}

\usepackage{xspace}
\newcommand{\adaQN}{\textsc{adaQN}\xspace}
\newcommand{\ADAM}{\textsc{Adam}\xspace}
\newcommand{\ADAGRAD}{\textsc{Adagrad}\xspace}
\newcommand{\RMSPROP}{\textsc{RMSProp}\xspace}

\begin{document}
\title{\adaQN: An Adaptive Quasi-Newton Algorithm for Training RNNs}

\author{Nitish Shirish Keskar\thanks{Department of Industrial Engineering and Management Sciences, Northwestern University, 
       Evanston, IL, USA. } 
        \and 
        Albert S. Berahas\thanks{Department of Engineering Sciences and Applied Mathematics, Northwestern University 
       Evanston, IL, USA.}
            }

\maketitle

\begin{abstract}{Recurrent Neural Networks (RNNs) are powerful models that achieve exceptional performance on a plethora pattern recognition problems. However, the training of RNNs is a computationally difficult task owing to the well-known ``vanishing/exploding'' gradient problem. Algorithms proposed for training RNNs either exploit no (or limited) curvature information and have cheap per-iteration complexity, or attempt to gain significant curvature information at the cost of increased per-iteration cost. The former set includes diagonally-scaled first-order methods such as \ADAGRAD and \ADAM, while the latter consists of second-order algorithms like Hessian-Free Newton and K-FAC. In this paper, we present \adaQN, a stochastic quasi-Newton algorithm for training RNNs. Our approach retains a low per-iteration cost while allowing for non-diagonal scaling through a stochastic L-BFGS updating scheme. The method uses a novel L-BFGS scaling initialization scheme and is judicious in storing and retaining L-BFGS curvature pairs. We present numerical experiments on two language modeling tasks and show that \adaQN is competitive with popular RNN training algorithms.}
\end{abstract}

\section{Introduction}

Recurrent Neural Networks (RNNs) have emerged as one of the most powerful tools for modeling sequences \cite{sutskever2013training,graves2012supervised}. They are extensively used in a wide variety of applications including language modeling, speech recognition, machine translation and computer vision \cite{graves2013speech,robinson1996use,bengio2003neural,mikolov2011rnnlm,graves2009offline}. RNNs are similar to the popular Feed-Forward Networks (FFNs), but unlike FFNs, allow for cyclical connectivity in the nodes. This enables them to have exceptional expressive ability, permitting them to model highly complex sequences. This expressiveness, however, comes at the cost of training difficulty, especially in the presence of long-term dependencies \cite{pascanu2012difficulty, bengio1994learning}. This difficulty, commonly termed as the ``vanishing/exploding'' gradient problem, arises due to the recursive nature of the network. Depending on the eigenvalues of the hidden-to-hidden node connection matrix during the Back Propagation Through Time (BPTT) algorithm, the errors either get recursively amplified or diminished making the training problem highly ill-conditioned. Consequently, this issue precludes the use of methods which are unaware of the curvature of the problem, such as Stochastic Gradient Descent (SGD), for RNN training tasks.

Many attempts have been made to address the problem of training RNNs. Some propose the use of alternate architectures; for e.g. Gated Recurrent Units (GRUs) \cite{cho2014learning} and Long Short-Term Memory (LSTM) \cite{hochreiter1997long} models. These network architectures do not suffer as severely from gradient-related problems, and hence, it is possible to use simple and well-studied methods like SGD for training, thus obviating the need for more sophisticated methods. Other efforts for alleviating the problem of training RNNs have been centered around designing training algorithms which incorporate curvature information in some form; see for e.g. Hessian-Free Newton \cite{martens2011learning,martens2010deep} and Nesterov Accelerated Gradient \cite{nesterov1983method}.

First-order methods such as \ADAGRAD \cite{duchi2011adaptive} and \ADAM \cite{kingma2014adam}, employ diagonal scaling of the gradients and consequently achieve invariance to diagonal re-scaling of the gradients. These methods have low per-iteration cost and have demonstrated good performance on a large number of deep learning tasks. Second-order methods like Hessian-Free Newton \cite{martens2010deep} and K-FAC \cite{martens2015optimizing}, allow for non-diagonal-scaling of the gradients using highly expressive Hessian information, but tend to either have higher per-iteration costs or require non-trivial information about the structure of the graph. We defer the discussion of these algorithms to the following section. 

In this paper, we present \adaQN, a novel (stochastic) quasi-Newton algorithm for training RNNs. The algorithm attempts to reap the merits of both first- and second-order methods by judiciously incorporating curvature information while retaining a low per-iteration cost. Our algorithmic framework is inspired by that of Stochastic Quasi-Newton (SQN) \cite{byrd2014stochastic}, which is designed for stochastic convex problems. The proposed algorithm is designed to ensure practical viability for solving RNN training problems. 

The paper is organized as follows. We end the introduction by establishing notation that will be used throughout the paper. In Section \ref{sec:rel_work}, we discuss popular algorithms for training RNNs and also discuss stochastic quasi-Newton methods. In Section \ref{sec:adaQN}, we describe our proposed algorithm in detail and emphasize its distinguishing features. We present numerical results on language modeling tasks in Section \ref{sec:num_res}. Finally, we discuss possible extensions of this work and present concluding remarks in Sections \ref{sec:disc} and \ref{sec:conc} respectively. 

\subsection{Notation}
The problem of training RNNs can be stated as the following optimization problem, 
\begin{align}
\label{eqn:prob}
   \min_{w\in\mathbb{R}^n}f(w)=\frac{1}{m}\sum_{i=1}^{m} f_{i}(w).
\end{align}
Here, $f_i$ is the RNN training error corresponding to a data point denoted by index $i$. We assume there are $m$ data points. During each iteration, the algorithm samples data points $\mathcal{B}_k \subseteq \{1,2,\cdots,m\}$. The iterate at the $k^{th}$ iteration is denoted by $w_k$ and the (stochastic) gradient computed on this mini-batch is denoted by $\hat{\nabla}_{\mathcal{B}_k} f(w_k)$. In particular, the notation $\hat{\nabla}_{\mathcal{B}_j} f(w_k)$ can be verbally stated as the gradient computed at $w_k$ using the mini-batch used for gradient computation during iteration $j$. For ease of notation, we may eliminate the subscript $\mathcal{B}_k$ whenever the batch and the point of gradient evaluation correspond to the iterate index; in other words, we use $\hat{\nabla} f(w_k)$ to mean $\hat{\nabla}_{\mathcal{B}_k} f(w_k)$. Unless otherwise specified, $H_k$ denotes any positive-definite matrix and the step-length is denoted by $\alpha_k$. The positive-definiteness of a matrix $H$ is expressed using the notation $H \succ 0$. Lastly, we denote the $i^{th}$ component of a vector $v\in\mathbb{R}^n$ by $[v]_i$ and use $v^2$ to represent element-wise square.

\section{Related Work}
\label{sec:rel_work}

In this section, we discuss several methods that have been proposed for training RNNs. In its most general form, the update equation for these methods can be expressed as
\begin{equation}  \label{genupd}
   w_{k+1}=w_{k} - \alpha_{k}H_{k}(\hat{\nabla}  f(w_k) + v_kp_k)
\end{equation} 
where $\hat{\nabla} f(w_k)$ is a stochastic gradient computed using batch $\mathcal{B}_k$; $H_k$ is a positive-definite matrix representing an approximation to the {inverse}-Hessian matrix; $p_k$ is a search direction (usually $w_{k}-w_{k-1}$) associated with a momentum term; and $v_k\geq0$ is the relative scaling of the direction $p_k$.

\subsection{Stochastic First-Order Methods}
Inarguably, the simplest stochastic first-order method is SGD whose updates can be represented in the form of \eqref{genupd} by setting $H_k=I$ and $v_k,p_k=0$. Momentum-based variants of SGD (such as Nesterov Accelerated Gradient \cite{nesterov1983method}) use $p_k=(w_{k}-w_{k-1})$ with a tuned value of $v_k$. While SGD has demonstrated superior performance on a multitude of neural network training problems \cite{bengio2015deep}, in the specific case of RNN training, SGD has failed to stay competitive owing to the ``vanishing/exploding'' gradients problem \cite{pascanu2012difficulty,bengio1994learning}.

 There are diagonally-scaled first-order algorithms that perform well on the RNN training task. These algorithms can  be interpreted as attempts to devise second-order methods via inexpensive diagonal Hessian approximations. \ADAGRAD  \cite{duchi2011adaptive} allows for the independent scaling of each  variable, thus partly addressing the issues arising from ill-conditioning. \ADAGRAD can be written in the general updating form by setting $v_k,p_k=0$ and by updating $H_k$ (which is a diagonal matrix) as
\begin{align*}
   [H_k]_{ii} &= \frac{1}{\sqrt{\sum_{j=0}^k[\hat{\nabla} f(w_j)]_i^2 + \epsilon}},
            \end{align*}
            where
            $\epsilon>0$ is used to prevent numerical instability arising from dividing by small quantities.
            
Another first-order stochastic method that is known to perform well empirically in RNN training is \ADAM \cite{kingma2014adam}. The update, which is a combination of \RMSPROP \cite{tieleman2012lecture} and momentum, can be represented as follows in the form of \eqref{genupd}, 
\begin{align*}
v_k &= \beta_1, &
    p_k &= \sum_{j=0}^{k-1} \beta_1^{(k-j-1)}(1-\beta_1) \hat{\nabla} f(w_j) - \hat{\nabla} f(w_k),\\
    r_k &= \sum_{j=0}^k \beta_2^{(k-j)}(1-\beta_2) \hat{\nabla} f(w_j)^2, & 
    [H_k]_{ii} &=   
                \frac{1}{\sqrt{[r_k]_i+\epsilon}}. 
\end{align*}

The diagonal scaling of the gradient elements in \ADAGRAD and \ADAM allows for infrequently occurring features (with low gradient components) to have larger step-sizes in order to be effectively learned, at a rate comparable to that of frequently occurring features. This causes the iterate updates to be more stable by controlling the effect of large (in magnitude) gradient components, to some extent reducing the problem of ``vanishing/exploding'' gradients.  However, these methods are not completely immune to curvature problems. This is especially true when the eigenvectors of $\nabla^2 f(w_k)$ do not align with the co-ordinate axes. In this case, the zig-zagging (or bouncing) behavior commonly observed for SGD may occur even for methods like \ADAGRAD and \ADAM. 


\subsection{Stochastic Second-Order Methods}
Let us first consider the Hessian-Free Newton methods (HF) proposed in \cite{martens2010deep, martens2011learning}. These methods can be represented in the form of \eqref{genupd} by setting $H_k$ to be an approximation to the inverse of the Hessian matrix ($\nabla^2 f(w_k)$), as described below, with the circumstantial use of momentum to improve convergence. HF is a second-order optimization method that has two major ingredients: (i) it implicitly creates and solves quadratic models using matrix-vector products with the Gauss-Newton matrix obtained using the ``Pearlmutter trick'' and (ii) it uses the Conjugate Gradient method (CG) for solving the sub-problems inexactly. Recently, \cite{martens2015optimizing}  proposed K-FAC, a method that computes a second-order step by constructing an invertible approximation of a neural networks' Fisher information matrix in an online fashion. The authors claim that the increased quality of the step offsets the increase in the per-iteration cost of the algorithm.

Our algorithm \adaQN belongs to the class of stochastic quasi-Newton methods which use a non-diagonal scaling of the gradient, while retaining low per-iteration cost. We begin by briefly surveying past work in this class of methods. 

\subsection{Stochastic Quasi-Newton Methods}

Recently, several stochastic quasi-Newton algorithms have been developed for large-scale machine learning problems: oLBFGS \cite{schraudolph2007stochastic,mokhtari2014global}, RES \cite{mokhtari2014res}, SDBFGS \cite{wang2014stochastic}, SFO \cite{sohl2013fast} and  SQN \cite{byrd2014stochastic}. These methods can be represented in the form of \eqref{genupd} by setting $v_k,p_k=0$ and using a quasi-Newton approximation for the matrix $H_k$. The methods enumerated above differ in three major aspects: (i) the update rule for the curvature pairs used in the computation of the quasi-Newton matrix, (ii) the frequency of updating, and (iii) the applicability to non-convex problems. With the exception of SDBFGS, all aforementioned methods have been designed to solve convex optimization problems. In all these methods, careful attention must be taken to monitor the quality of the curvature information that is used. 

The RES and SDBFGS algorithms control the quality of the steps by modifying the BFGS update rule \cite{nocedal2006numerical}. Specifically, the update equations take on the following form,
\begin{align}
s_k &= w_{k+1} - w_{k},\\
y_k &= \hat{\nabla}_{\mathcal{B}_k} f(w_{k+1}) - \hat{\nabla}_{\mathcal{B}_k} f(w_k) - \delta s_k  \label{eqn:res_y},\\
H_{k+1}^{-1} = B_{k+1} &= B_k + \frac{y_k^Ty_k}{y_k^Ts_k} - \frac{B_k s_k s_k^T B_k}{s_k^T B_k s_k} + \delta I.
\end{align}
This ensures that the Hessian approximations are uniformly bounded away from singularity, thus preventing the steps from becoming arbitrarily large. Further, in these methods, the line-search is replaced by a decaying step-size rule. Note that at the $k^{th}$ iteration, the gradients used during updates \eqref{eqn:res_y} are both evaluated on $\mathcal{B}_k$. oLBFGS, is similar to the above methods except no $\delta$-modification is used. In the equations above, $B_k$ and $H_k$ denote approximations to the Hessian and inverse-Hessian matrices respectively. 



Finally, in \cite{byrd2014stochastic}, the authors propose a novel quasi-Newton framework, SQN, in which they recommend the decoupling of the stochastic gradient calculation from the curvature estimate. The BFGS matrix is updated once every $L$ iterations as opposed to every iteration, which is in contrast to other methods described above. The authors prescribe the following curvature pair updates,
\begin{align}  
        s_t &= \bar{w}_{t} - \bar{w}_{t-1}, & \text{where } \bar{w}_t = \frac{1}{L} \sum_{i=(t-1)L}^{tL} w_i,\label{eq:curvSQN_s}\\
        y_t &= \hat{\nabla}^2_{\mathcal{H}_t}F(\bar{w}_t)s_t, & \label{eq:curvSQN_y}
\end{align}
where $t$ is the curvature pair update counter, $L$ is the update frequency (also called the aggregation length) and $\mathcal{H}_t$ is a mini-batch used for computing the sub-sampled Hessian matrix. The iterate difference, $s$, is based on the average of the iterates over the last $2L$ iterations, intuitively allowing for more stable approximations. On the other hand, the gradient differences, $y$, are not computed using gradients at all, rather they are computed using a Hessian-vector product representing the approximate curvature along the direction $s$. 

The structure of the curvature pair updates proposed in SQN has several appealing features. Firstly, updating curvature information, and thus the Hessian approximation, every $L$ iterations (where $L$ is typically between $2$ and $20$) considerably reduces the computational cost. Additionally, more computational effort can be expended for the curvature computation since this cost is amortized over $L$ iterations. Further, as explained in \cite{byrd2014stochastic}, the use of the Hessian-vector product in lieu of gradient differences allows for a more robust estimation of the curvature, especially in cases when $\|s\|$ is small and the gradients are noisy. 

The SQN algorithm was designed specifically for convex optimization problems arising in machine learning, and its extension to RNN training is not trivial. In the following section, we describe \adaQN, our proposed algorithm, which uses the algorithmic framework of SQN as a foundation. More specifically, it retains the ability to decouple the iterate and update cycles along with the associated benefit of investing more effort in gaining curvature information.

\section{\adaQN}
\label{sec:adaQN}

In this section, we describe the proposed algorithm in detail. Specifically, we address key ingredients of the algorithm, including (i) the initial L-BFGS scaling, (ii) step quality control, (iii) choice of Hessian matrix for curvature pair computation, and (iv) the suggested choice of hyper-parameters. The pseudo-code for \adaQN is given in Algorithm~\ref{alg:adaQN}.

\begin{algorithm}[H]
\caption{\adaQN}
\label{alg:adaQN}
{\bf Inputs:} $w_0$,  $L$, $\alpha$, sequence of batches $\mathcal{B}_k$ with $|\mathcal{B}_k|=b$ for all $k$, $m_L=10$, $m_F=100$, $\epsilon=10^{-4}$, $\gamma=1.01$
\smallskip{}
\begin{algorithmic}[1]
\State Set $t \leftarrow 0$ and $\bar{w}_{o}, w_s=0$
\State Initialize accumulated Fisher Information matrix FIFO container $\tilde{F}$ of maximum size $m_F$ and L-BFGS curvature pair containers $S,Y$ of maximum size $m_L$. 
\State Randomly choose a mini-batch as monitoring set $\mathcal{M}$

\For {$k=0,1,2,...$}

        \State $w_{k+1} = w_{k} - \alpha H_k \hat{ \nabla}f(w_k)$ \Comment{Compute adaQN updates using two-loop recursion}
        \State Store $\hat{ \nabla} f(w_k) \hat{ \nabla} f(w_k)^T$ in $\tilde{F}$

    \State $w_s = w_s + w_{k+1}$                \Comment{Running sum of iterates for average computation}

    \If {$\mod(k,L)=0$}
        
        \State $\bar{w}_n = \frac{w_s}{L}$          \Comment{Compute average iterate}
        \State $w_s = 0$                            \Comment{Clear accumulated sum of iterates}
        
        \If {$t>0$}

        \If{$f_\mathcal{M}(\bar{w}_n) > \gamma f_\mathcal{M}(\bar{w}_o)$}       \Comment{Check for step rejection}
        
            \State Clear L-BFGS memory and the accumulated Fisher Information container $\tilde{F}$.
            \State $w_k = \bar{w}_{o}$              \Comment{Return to previous aggregated point}
            \State \textbf{continue}
            
        \EndIf
        
        \State $s = \bar{w}_n - \bar{w}_{o}$              \Comment{Compute curvature pair}
        \State $y = \frac{1}{|\tilde{F}|}(\sum_{i=1}^{|\tilde{F}|} \tilde{F}_i \cdot s)$      \Comment{Compute curvature pair}
        
        \If {${s^Ty}>\epsilon \cdot {s^Ts}$}          \Comment{Check for sufficient curvature}
            \State Store curvature pairs $s_t$ and $y_t$ in containers $S$ and $Y$ respectively
            \State $\bar{w}_{o} = \bar{w}_n$
        \EndIf
        
        \Else
        \State $\bar{w}_{o} = \bar{w}_n$
        \EndIf
    
    \State $t \gets t + 1$
        
    \EndIf

\EndFor

\end{algorithmic}
\end{algorithm}

We emphasize that the storage of $(\hat{ \nabla} f(w_k)\hat{ \nabla} f(w_k)^T)$ in Step 6 is for ease of notation; in practice it is sufficient to store $\hat{ \nabla} f(w_k)$ and compute $y$ in Step 18 without explicitly constructing the matrix. Also, the search direction $p_k=-H_k\hat{\nabla}f_k$ is computed via the two-loop recursion using the available curvature pairs $(S,Y)$, and thus the matrix $H_k$ (the approximation to the inverse-Hessian matrix) is never constructed; refer to Section \ref{scn:two-loop}. Further, in Algorithm \ref{alg:adaQN}, we specify a fixed monitoring set $\mathcal{M}$, a feature of the algorithm that was set for ease of exposition. In practice, this set can be changed 
to allow for lower bias in the step acceptance criterion.

\subsection{Choice of $H^{(0)}_k$ for L-BFGS}
\label{scn:hk0}

Firstly, we discuss the most important ingredient of the proposed algorithm: the initial scaling of the L-BFGS matrix. For L-BFGS, in both the deterministic and stochastic settings, a matrix $H^{(0)}_k$ must be provided, which is an estimate of the scale of the problem. This choice is crucial since the relative scale of the step (in each direction) is directly related to it.
In deterministic optimization, \begin{align} H^{(0)}_k &= \frac{s_k^T y_k}{y_k^Ty_k} I \label{eqn:bb_scaling} \end{align} is found to work well on a wide variety of applications, and is often prescribed \cite{nocedal2006numerical}. Stochastic variants of L-BFGS, including oBFGS, RES and SQN, prescribe the use of this initialization (\eqref{eqn:bb_scaling}). However, this is dissatisfying in the context of RNN training for two reasons. Firstly, as mentioned in the previous sections, the issue of  ``vanishing/exploding'' gradients makes the problems highly ill-conditioned; using a scalar initialization of the L-BFGS matrix does not address this issue. Secondly, since $s$ and $y$ are noisy estimates of the true iterate and gradient differences, the scaling suggested in \eqref{eqn:bb_scaling} could introduce adversarial scale to the problem, causing performance deterioration.

To counter these problems, we suggest an initialization of the inverse-Hessian matrix based on accumulated gradient information. Specifically, we set
\begin{equation}  \label{adaQN_scale}
   [H_k^{(0)}]_{ii} = \frac{1}{\sqrt{\sum_{j=0}^k[\hat{\nabla} f(w_j)]_i^2 +\epsilon}}, \forall i={1,...,n}.
\end{equation} 
We direct the reader to Section \ref{scn:two-loop} for details on how the above initialization is used as part of the L-BFGS two-loop recursion.
We emphasize that this initialization is: (i) a diagonal matrix with non-constant diagonal entries, (ii) has a cost comparable to \eqref{eqn:bb_scaling}, and (iii) is identical to the scaling matrix used by \ADAGRAD at each iteration. This choice is motivated by our observation of \ADAGRAD's stable performance on many RNN learning tasks. By initializing L-BFGS with an \ADAGRAD-like scaling matrix, we impart a better scale in the L-BFGS matrix, and also allow for implicit safeguarding of the proposed method. Indeed, in iterations where no curvature pairs are stored, the \adaQN and \ADAGRAD steps are identical in form.  

\subsection{Step Acceptance and Control}
While curvature information can be used to improve convergence rates, noisy or stale curvature information may in fact deteriorate performance \cite{byrd2014stochastic}. SQN attempts to prevent this problem by using large-batch Hessian-vector products in \eqref{eq:curvSQN_y}. Other methods attempt to control the quality of the steps by modifying the L-BFGS update rule to ensure that $H_k \succ 0$ for all $k$. However, we have found that these do not work well in practice. Instead, we control the quality of the steps by judiciously choosing the curvature pairs used by L-BFGS. We attempt to store curvature pairs during each cycle but skip the updating if the calculated curvature is small; see \cite{nocedal2006numerical} for details regarding skipping in quasi-Newton methods. Further, we flush the memory when the step quality deteriorates, allowing for more reliable steps till the memory builds up again.


The proposed criterion (Line 12 of Algorithm \ref{alg:adaQN}) is an inexpensive heuristic wherein the functions are evaluated on a monitoring set, and $\gamma$ approximates the effect of noise on the function evaluations. A step is rejected if the function value of the new aggregated point is significantly worse (measured by $\gamma$) than the previous. In this case, we reset the memory of L-BFGS which allows the algorithm to preclude the deteriorating effect of any stored curvature pairs. The algorithm resumes to take \ADAGRAD steps and build up the curvature estimate again. We report that, as an alternative to the proposed criterion, a more sophisticated criterion such as \textit{relative improvement}, \begin{align*}
\frac{f_\mathcal{M}(\bar{w}_n)-f_\mathcal{M}(\bar{w}_o)}{f_\mathcal{M}(\bar{w}_o)} &> \tilde{\gamma} \in (0,1)
\end{align*}
delivered similar performance on our test problems.

In the case when the sufficient curvature condition (Line 19 of Algorithm \ref{alg:adaQN}) is not satisfied, the storage of the curvature pair is skipped. In deterministic optimization, this problem is avoided by conducting a Wolfe line-search. If the curvature information for a given step is inadequate, the line-search attempts to look for points further along the search path. We extend this idea to the RNN setting by not updating $\bar{w}_{o}$ when this happens. This allows us to move further, and possibly glean curvature information in  subsequent update attempts. We have experimentally found this safeguarding to be crucial for the robust performance of \adaQN. That being said, such rejection happens infrequently and the average L-BFGS memory per epoch remains high for all of our reported experiments (see Section~\ref{sec:num_res}).

\subsection{Choice of Curvature Information Matrix}
\label{scn:curv}
As in SQN, the iterate difference $s$ in our algorithm is computed using aggregated iterates and the gradient difference $y$ is computed through a matrix-vector product; refer to equations \eqref{eq:curvSQN_s} and \eqref{eq:curvSQN_y}. The choice of curvature matrix for the computation of $y$ must address the trade-off between obtaining informative curvature information and the computational expense of its acquisition. 
Recent work suggests that the Fisher Information matrix (FIM) yields a better estimate of the curvature of the problem as compared to the true Hessian matrix (which is a natural choice); see for e.g. \cite{martens2014new}.

Given a function $f$ parametrized by a random variable $\mathcal{X}$, the (true) FIM at a point $w$ is given by
\begin{align*}
F(w) &= \mathbb{E}_{\mathcal{X}} [\nabla f_{\mathcal{X}}(w) \nabla f_{\mathcal{X}}(w)^T].
\end{align*}
Since the distribution for $\mathcal{X}$ is almost never known, the \textit{empirical} Fisher Information matrix (eFIM) is often used in practice. The eFIM can be expressed as follows
\begin{align}
\label{eq:emp_fish}
\hat{F}(w) &= \frac{1}{|\mathcal{H}|} \sum_{i\in\mathcal{H}} \nabla_i f(w) \nabla_i f(w)^T,
\end{align}
where $\mathcal{H} \subseteq \{1,2,\cdots,m\}$.

Notice from equation~\eqref{eq:emp_fish} that the eFIM is guaranteed to be positive semi-definite, a property that does not hold for the true Hessian matrix. The use of the FIM (or eFIM) in second-order methods allows for attractive theoretical and practical properties. We exclude these results for brevity and refer the reader to \cite{martens2014new} for a detailed survey regarding this topic. 

 
 Given these observations and results, the use of the eFIM may seem like a reasonable choice for the Hessian matrix approximation used in the computation of $y_t$ (see equation \eqref{eq:curvSQN_y}). However, the use of this matrix, even infrequently, increases the amortized per-iteration cost as compared to state-of-the-art first-order stochastic methods.
 Further, unlike second-order methods which rely on relatively accurate curvature information to generate good steps, quasi-Newton methods are able to generate high-quality steps even with crude curvature information \cite{nocedal2006numerical}. In this direction, we propose the use of a modified version of the empirical Fisher Information matrix that uses historical values of stochastic gradients, which were already computed as part of the step, thus reducing the computational cost considerably. This reduction, comes at the expense of storage and potentially noisy estimates due to stale gradient approximations. We call this approximation of the eFIM the \textit{accumulated} Fisher Information matrix (aFIM) and denote it by $\bar{F}$. Given a memory budget of $m_F$, the aFIM at the $k^{th}$ iteration is given by
\begin{align}  
\label{eq:acc_fish}
   \bar{F}(w_k) &= \frac{1}{\sum_{j=k-m_F+1}^k \left| \mathcal{B}_j \right|} \sum_{j=k-m_F+1}^k \nabla_{\mathcal{B}_j} f(w_{j})\nabla_{\mathcal{B}_j} f(w_{j})^T.
\end{align}

For the purpose of our implementation, we maintain a finite-length FIFO container $\tilde{F}$ for storing the stochastic gradients as they are computed. Whenever the algorithm enters lines 12--16, we reject the step, and the contents of $\tilde{F}$ along with the L-BFGS memory are cleared. By clearing $\tilde{F}$, we also allow for additional safeguarding of future iterates against noisy gradients in the $\tilde{F}$ container that may have contributed in the generation of the poor step. 






\subsection{Choice of Hyper-Parameters}

\adaQN has a set of hyper-parameters that require tuning for competitive performance. Other than the step-size and batch-size, which needs to be tuned for all aforementioned methods, the only hyper-parameter exposed to the user is $L$. We prescribe $L$ to be chosen from $\{2,5,10,20\}$. We experimentally observed that the performance was not highly sensitive to the choice of $\alpha$ and $L$. Often, $L=5$ and the same step-length as used for \ADAGRAD gave desirable performance. The other hyper-parameters have intuitive default values which we have found to work well for a variety of applications. Additional details about the offline tuning costs of \adaQN as compared to \ADAGRAD and \ADAM can be found in Section \ref{sec:num_res}.

\subsection{Cost}
\label{scn:cost}
Given the nature of the proposed algorithm, a reasonable question is about the per-iteration cost. Let us begin by first considering the per-iteration cost of other popular methods. For simplicity, we assume that the cost of the gradient computation is $\mathcal{O}(n)$, which is a reasonable assumption in the context of deep learning. SGD has one of the cheapest per-iteration costs, with the only significant expense being the computation of the mini-batch stochastic gradient. Thus, SGD has a per-iteration complexity of $\mathcal{O}(n)$. \ADAGRAD and \ADAM also have the same per-iteration complexity since the auxiliary operations only involve dot-products and elementary vector operations. Further, these algorithms have $\mathcal{O}(1)$ space complexity. On the other hand, second-order methods have higher per-iteration complexity since each iteration requires an inexact solution of a linear system, and possibly, storage of the pre-conditioning matrices. 

The per-iteration time complexity of our algorithm consists of three components: (i) the cost of gradient computation, (ii) the cost of the L-BFGS two-loop recursion, and (iii) the amortized cost of computing the curvature pair. Thus, the overall cost can be written as~\begin{align}
\label{eqn:complexity}
\underbrace{\mathcal{O}(n)}_{\text{gradient computation}} + \underbrace{4m_Ln}_{\text{two-loop recursion}} + \underbrace{m_FnL^{-1}}_{\text{cost of computing curvature pair}}.
\end{align}
Given the prescription of $L\approx 5$, $m_F=100$ and $m_L=10$, the cost per-iteration remains at $\mathcal{O}(n)$. The memory requirement of our algorithm is also $\mathcal{O}(n)$ since we require the storage of up to $m_F + 2m_L$ vectors of size $n$. 

This result is similar to the one presented in \cite{byrd2014stochastic}. The difference in the complexity arises in the third term of \eqref{eqn:complexity} due to our choice of the accumulated Fisher Information matrix as opposed to using a sub-sampled Hessian approximation. It is not imperative for our algorithm to use aFIM for the computation of $y_t$ \eqref{eq:curvSQN_y}. We can instead use the eFIM \eqref{eq:emp_fish}, which would allow for a lower memory requirement (from $(m_F+m_L)n$ to $m_Ln$) at the expense of added computation during curvature pair estimation. However, the time complexity would remain linear in $n$ for either choice. As we mention in Section \ref{scn:curv}, by using the accumulated Fisher Information matrix, we avoid the need for additional computation at the expense of memory; a choice we have found to work well in practice.

\section{Numerical Results}
\label{sec:num_res}
In this section, we present numerical evidence demonstrating the viability of the proposed algorithm for training RNNs. We also present meta-data regarding the experiments which suggests that the performance difference between \adaQN and its competitors (and \ADAGRAD in particular) can be attributed primarily to the incorporation of curvature.

\subsection{Language Modeling}
\label{scn:numerical_lm}
For benchmarking, we compared the performance of \adaQN against \ADAGRAD and \ADAM on two  language modeling (LM) tasks: character-level LM and word-level LM. For the character-level LM task \cite{karpathy2015visualizing}, we report results on two data sets: The Tale of Two Cities (Dickens) and The Complete Works of Friedrich Nietzsche (Nietzsche) . The former has 792k characters while the latter has 600k. We used the Penn-Tree data set for the word-level LM task \cite{zaremba2014recurrent}. This data set consists of 929k training words with 10k
words in its vocabulary. 

\begin{figure}[H]
\centering
\includegraphics[width=0.75\textwidth]{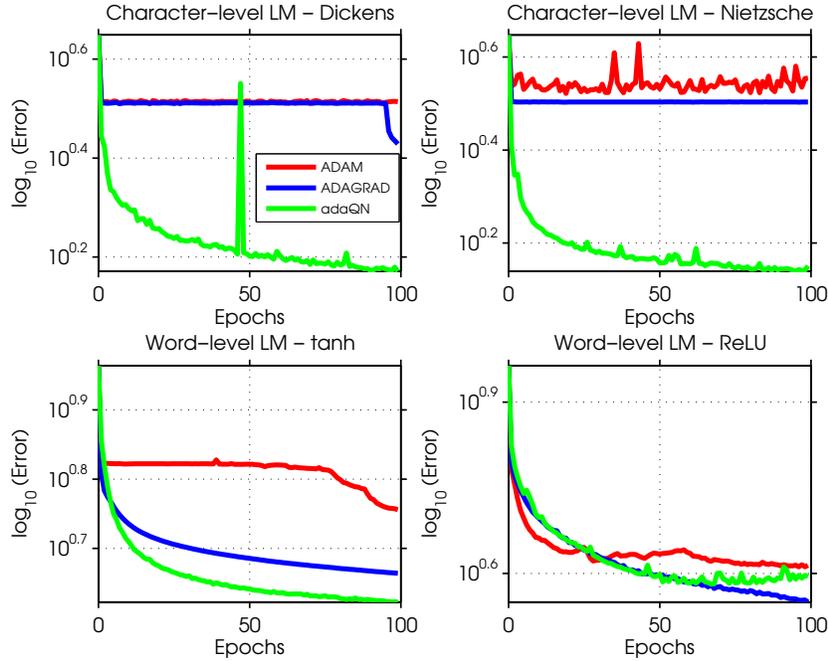}
\caption{Numerical Results on LM Tasks}
\label{fig:performance}
\end{figure}

For all tasks, we used an RNN with 5 recurrent layers. The input and output layer sizes were determined by the vocabulary of the data set. The character-level and word-level LMs were constructed with 100 and 400 nodes per layer respectively. The weights were randomly initialized from $\mathcal{N}(0,0.01)$. Unless otherwise specified, the activation function used was $\tanh$. The sequence length was chosen to be $50$ for both cases. For readability, we exclude other popular methods that did not consistently perform competitively. In particular, SGD (with or without momentum) was not found to be competitive despite significant tuning. For \adaQN, \ADAGRAD and \ADAM, all hyper-parameters were set using a grid-search. In particular, step-sizes were tuned for all three methods. \ADAM needed coarse-tuning for $(\beta_1,\beta_2)$ in the vicinity of the suggested values. For \adaQN, the value of $L$ was chosen from $\{2,5,10,20\}$. The rest of the hyper-parameters ($m_F,m_L,\epsilon, \gamma$) were set at their recommended values for all experiments (refer to Algorithm \ref{alg:adaQN}). It can thus be seen that the offline tuning costs of \adaQN are comparable to those of \ADAGRAD and \ADAM. We ran all experiments for 100 epochs and present the results (testing error) in Figure \ref{fig:performance}.


\begin{figure}[H]
\centering
\includegraphics[width=0.75\textwidth]{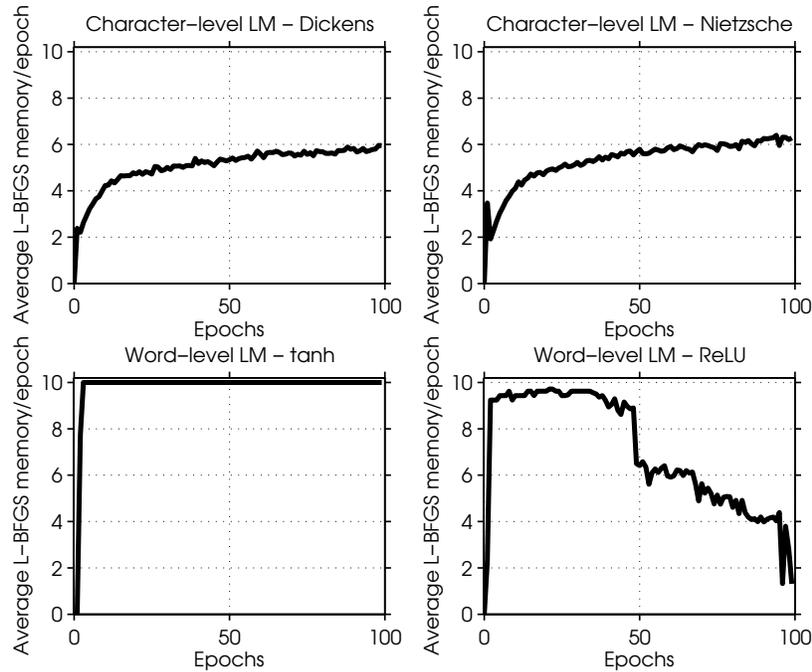}
\caption{Average L-BFGS Memory Per Epoch}
\label{fig:memory}
\end{figure}

It is clear from Figure \ref{fig:performance} that \adaQN presents a non-trivial improvement over both \ADAGRAD and \ADAM on all tasks with $\tanh$ activation function. Specifically, we emphasize the performance gain over \ADAGRAD, the method which \adaQN is safeguarded by. On the character-level task with ReLU activation, \adaQN performed better than \ADAM but worse than \ADAGRAD.  We point out that experiments with other (including larger) data sets yielded results of similar nature.

\subsection{Average L-BFGS Memory per Epoch}
\label{scn:avg_memory}
Given the safeguarded nature of our algorithm, a natural question regarding the numerical results presented pertains to the effect of the safeguarding on the performance of the algorithm. To answer this question, we report the average L-BFGS memory per epoch in Figure \ref{fig:memory}. This is computed by a running sum initialized at $0$ at the start of each new epoch. A value greater than $1$ indicates that at least one curvature pair was present in the memory (in expectation) during a given epoch. Higher average values of L-BFGS memory suggest that more directions of curvature were successfully explored; thus, the safeguarding was less necessary. Lower values, on the other hand, suggest that the curvature information was either not informative (leading to skipping) or led to deterioration of performance (leading to step rejection).

The word-level LM task with the ReLU activation function has interesting outcomes. It can be seen from Figure \ref{fig:performance} that the performance of \ADAGRAD is similar to that of \adaQN for the first 50 epochs but then \ADAGRAD continues to make progress while the performance of \adaQN stagnates. During the same time, the average L-BFGS memory drops significantly suggesting that safeguarding was necessary and that, the curvature information was not informative enough and even caused deterioration in performance (evidenced by occasional increase in the function value). 

\subsection{MNIST Classification from Pixel Sequence}
\label{scn:mnist}
\begin{figure}[H]
 \centering
 \includegraphics[width=0.75\textwidth]{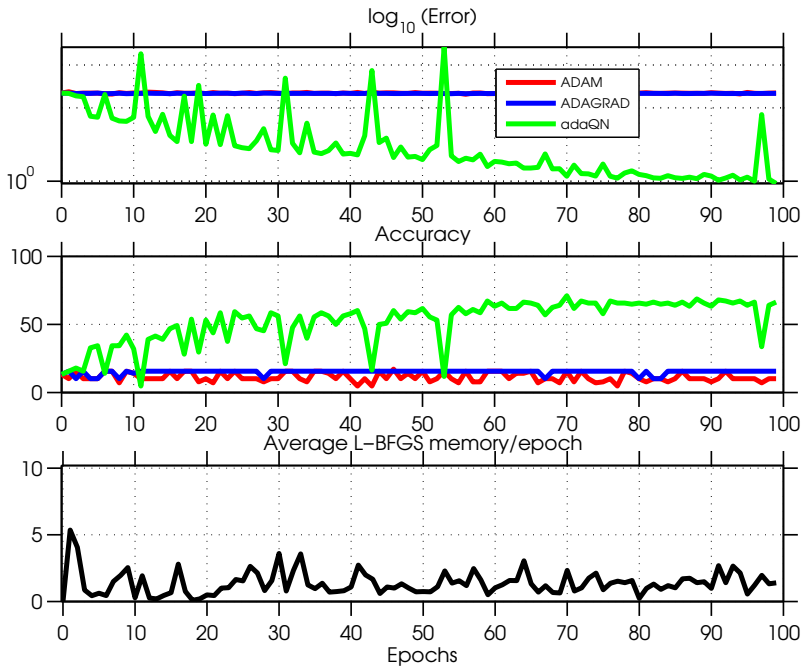}
 \caption{Numerical Results on MNIST with Sequence of Pixels}
 \label{fig:irnn}
 \end{figure}

A challenging toy problem for RNNs is that of image classification given pixel sequences \cite{le2015simple}. For this problem, the image pixels are presented sequentially to the network one-at-a-time and the network must predict the corresponding category. This long range dependency makes the RNN difficult to train. We report results for the popular MNIST data set. For this experiment, we used a setup similar to that of \cite{le2015simple} with two modifications: we used $\tanh$ activation function instead of ReLU and initialized all weights from $\mathcal{N}(0,0.01)$ instead of using their initialization trick. The results are reported in Figure \ref{fig:irnn}.

As can be seen from the Figure \ref{fig:irnn}, \ADAM and \ADAGRAD struggle to make progress and stagnate at an error value close to that of the initial point. On the other hand, \adaQN is able to significantly improve the error values, and also achieves superior classification accuracy rates. Experiments on other toy problems with long range dependencies, such as the addition problem \cite{hochreiter1997long}, yielded similar results.

\subsection{LSTMs}
\begin{figure}[H]
 \centering
 \includegraphics[width=0.75\textwidth]{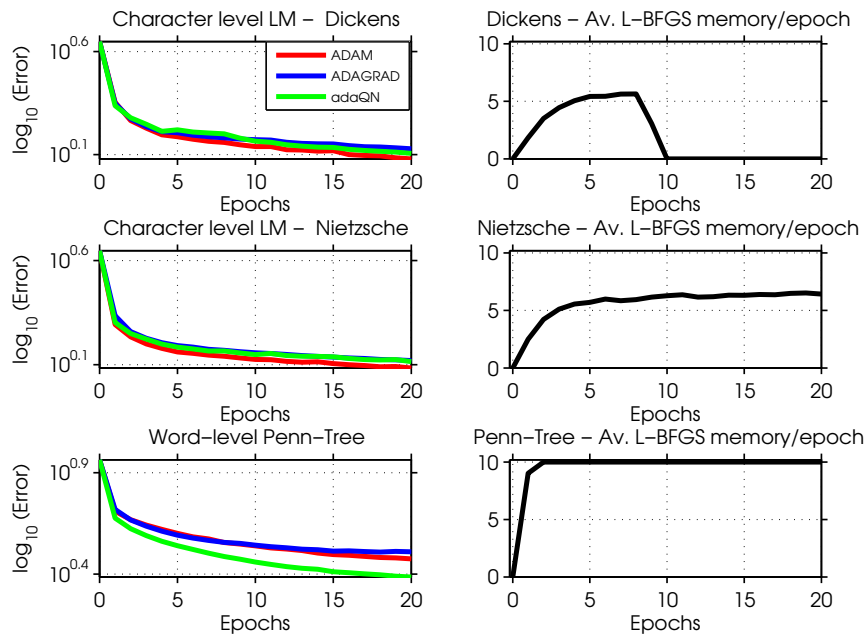}
 \caption{Numerical Results on LSTMs}
 \label{fig:LSTM}
 \end{figure}

In order to ascertain the viability of \adaQN on other architectures, we conducted additional experiments using the LSTM models. The experimental setup is similar to the one discussed in Section \ref{scn:numerical_lm} with the modification that 2 recurrent (LSTM) layers were used instead of 5. The results are reported in Figure \ref{fig:LSTM}.

 The results in Figure \ref{fig:LSTM} suggest mixed results. For the character-level LM tasks, the performance of \ADAGRAD and \adaQN was comparable while the performance of \ADAM was better. For the word-level LM task, the performance of \adaQN was superior to that of both \ADAGRAD and \ADAM.

\section{Discussion}
\label{sec:disc}
The results presented in the previous section suggest that \adaQN is competitive with popular algorithms for training RNNs. However, \adaQN is not restricted to this class of problems. Indeed, preliminary results on other architectures (such as Feed-Forward Networks) delivered promising performance. It may be possible to further improve the performance of the algorithm by modifying the update rule and frequency. In this direction, we discuss the practicality of using momentum in such an algorithm and possible heuristics to allow the algorithm to adapt the cycle length $L$ as opposed to tuning it to a constant value. 

Recent work by \cite{sutskever2013importance} suggests superior performance of momentum-based methods on a wide variety of learning tasks. These methods, with the right initialization, have been shown to outperform sophisticated methods such as the Hessian-Free Newton method. However, recent efforts suggest the use of second-order methods \textit{in conjunction} with momentum \cite{martens2015optimizing,martens2014new}. In this case, one interpretation of momentum is that of providing a  pre-conditioner to the CG sub-solver. Significant performance gains through the inclusion of momentum have been reported when the gradients are reliable \cite{martens2014new}. We hypothesize that performance gains can be obtained through careful inclusion of momentum for methods like \adaQN as well. However, the design of such an algorithm, and efficacy of using momentum-like ideas is an open question for future research. 

Lastly, we discuss the role of the aggregation cycle length $L$ on the performance of the algorithm. If $L$ is chosen to be too large, the aggregation points will be too far-apart possibly leading to incorrect curvature estimation. If $L$ is too small, then the iterates change insufficiently before an update attempt is made leading to skipping of update pairs. Besides the issue of curvature quality, the choice of $L$ also has ramifications on the cost of the algorithm as discussed in Section \ref{scn:cost}. Thus, a natural extension of \adaQN is an algorithm where $L$ can be allowed to adapt during the course of the algorithm. $L$ could be increased or decreased depending on the quality of the estimated curvature, while being bounded to ensure that the cost of updating is kept at a reasonable level. The removal of this hyper-parameter will not only obviate the need for tuning, but will also allow for a more robust performance.

\section{Conclusions}
\label{sec:conc}
In this paper, we present a novel quasi-Newton method, \adaQN, for training RNNs. The algorithm judiciously incorporates curvature information while retaining a low per-iteration cost. The algorithm builds upon the framework proposed in \cite{byrd2014stochastic}, which was designed for convex optimization problems. We discuss the key ingredients of our algorithm, such as, the scaling of the L-BFGS matrices using historical gradients,  curvature pair updating and step acceptance criterion, and, suggest the use of an accumulated Fisher Information matrix during the computation of a curvature pair. We examine the per-iteration time and space complexity of \adaQN and show that it is of the same order of magnitude as popular first-order methods. Finally, we present numerical results for two language modeling tasks and demonstrate competitive performance of \adaQN as compared to popular algorithms used for training RNNs.

\newpage{}
\small
\bibliographystyle{plain}
\bibliography{adaQN_refs}

\newpage{}

\appendix
\normalsize
\section{Appendix}
\subsection{L-BFGS Two-Loop Recursion}
\label{scn:two-loop}
We describe the two-loop recursion used to compute the step $p_k$ in Algorithm \ref{alg:two_loop}. We refer the reader to \cite{nocedal2006numerical} for additional details.

\begin{algorithm}[H]
\caption{Two Loop Recursion}
\label{alg:two_loop}
{\bf Inputs:} Curvature pair containers $S$ and $Y$, $\hat{\nabla} f(w_k)$

\begin{algorithmic}[1]

\State $\tau = \text{length}(S)$   \Comment{Compute the number of curvature pairs in memory}

\State $q \leftarrow \hat{ \nabla} f(w_k)$ 

\For {$i=\tau,\tau -1,...,1$}     \Comment{Backward Loop}

\State $\alpha_i\leftarrow \rho_i s_i^Tq$

\State $q \leftarrow q - \alpha_iy_i$

\EndFor

\State $r \leftarrow H_k^{(0)} q$

\For {$i=1,2,...,\tau$}    \Comment{Forward Loop}

\State $\beta \leftarrow \rho_i y_i ^Tr$

\State $r \leftarrow r + s_i(\alpha_i - \beta)$

\EndFor

\State $H_k \hat{\nabla} f(w_k) = r$

\end{algorithmic}

{\bf Output:} $p_k = -H_k \hat{\nabla} f(w_k)$    \Comment{Output the quasi-Newton search direction}
\end{algorithm}

L-BFGS is an overwriting process as opposed to an updating process; each curvature pair modifies $H_k^{(0)}$ via a rank-2 update. Thus, it is possible to reduce the effect of an incorrect estimation of $H^{(0)}_k$ through sufficient curvature pair updates. However, the initial estimate of scaling $H^{(0)}_k$ remains an important ingredient of L-BFGS updating, especially when the L-BFGS memory is low. If this value is incorrectly estimated, not only does it affect the scale of step, it also affects its quality. If the L-BFGS memory $\tau$ is low, it presents limited avenue for the algorithm to overwrite the poor scale imparted by $H^{(0)}_k$. Indeed, if $\tau$ is $0$, i.e. no curvature pairs are stored, the step that is returned by the two-loop recursion is $p_k = -H_k^{(0)} \hat{\nabla} f(w_k)$. This necessitates the choice of a good scaling matrix for $H_k^{(0)}$ in the case of stochastic quasi-Newton methods. As we mention in Section \ref{scn:hk0}, \adaQN uses an \ADAGRAD-like scaling matrix for $H_k^{(0)}$.

\end{document}